\documentclass[letterpaper, 10 pt, conference]{ieeeconf}

\usepackage{cite}
\usepackage{blindtext}
\usepackage{graphicx}
\usepackage{tabularx}
\usepackage{tabulary}
\usepackage{booktabs}
\usepackage{amsmath}
\usepackage{stackengine}
\usepackage{amssymb}
\usepackage{mathtools}
\usepackage{resizegather} 
\usepackage[utf8]{inputenc}
\usepackage{textcomp}
\usepackage{gensymb}
\usepackage{multirow}
\usepackage{multicol}
\usepackage[none]{hyphenat}
\usepackage{lipsum,multicol}
\usepackage{upgreek}
\usepackage{makecell}
\usepackage{array}
\usepackage[mathscr]{euscript}
\usepackage{algpseudocode}
\usepackage{xcolor}
\usepackage[hidelinks]{hyperref}
\usepackage{xurl}

\DeclarePairedDelimiterX\set[1]\lbrace\rbrace{\def\given{\;\delimsize\vert\;}#1}

\graphicspath{{Figures/}}

\IEEEoverridecommandlockouts                              
\title{\LARGE \bf Geometric Workspace Analysis and Transmission-Aware Dynamics of a Serial Spherical Tool for Microsurgery}

\author{Anestis~Mablekos-Alexiou,
        Lyndon~da~Cruz,
        and~Christos~Bergeles%
\thanks{This research was supported by the Sir Michael Uren Foundation.}%
\thanks{A.~Mablekos-Alexiou and L.~da~Cruz are with Moorfields Eye Hospital NHS Foundation Trust, London, UK. C. Bergeles is with King's College London. Correspondence: \texttt{mamplekos@gmail.com}.}
}

\begin{document}

\maketitle
\thispagestyle{empty}
\pagestyle{empty}

\begin{abstract}

We present a kinematic and transmission-aware design framework for a serial spherical mechanism with an additional translational degree of freedom for microsurgery. The first contribution is an analytical workspace formulation that provides geometric insight into reachable motion and enables rapid selection of rotation axis orientations without numerical optimization. The second contribution is a dynamics-informed methodology for mechanisms driven by self-locking transmissions, supporting evaluation of torque requirements for a prescribed workspace geometry. The framework is accompanied by an open-source software package for friction identification and inverse dynamics analysis. Experiments on a purpose-built robotic tool for vitreoretinal surgery validate the predictive capability of the models and demonstrate their practical utility for engineering design.

\end{abstract}


\section{Introduction}
\label{sec:intro}

Spherical mechanisms, whether in serial~\cite{hannaford,patent1} or parallel configurations~\cite{patent2}, consist of pin-jointed spatial linkages whose axes of revolution intersect at a single point in space. This property, which produces a remote center of motion (RCM), makes them highly compatible with robotic surgery as an alternative to parallelogram-based mechanisms.  

The literature on spherical mechanisms covers parallel \cite{ouerfelli,li,bai,degirmenci,essomba,herve,kong,zoppi1,zoppi2}, scissor \cite{afshar}, and serial configurations \cite{lum1,lum2,kim,laribi}. Specifically, for the serial spherical mechanism (SSM), Lum \textit{et al.} \cite{lum1,lum2} and Laribi \textit{et al.}~\cite{laribi} proposed a four degree of freedom (DoF) robot for minimally invasive surgery, while Kim \textit{et al.}~\cite{kim} presented a 5-DoF manipulator for robotic endoscopy. Overall, the majority of previous works on the SSM have focused on its kinematics, often exploiting numerical methods to optimize the manipulator structure. While these methods provide reliable solutions, they often offer limited geometric insight into the relationship between joint-axis configuration and achievable motion, making rapid analysis and early-stage design decisions less intuitive. We show that a geometric formulation leads to an analytically interpretable description of mechanism motion, providing insight into workspace structure while remaining directly applicable to analysis and engineering design. 

\begin{figure}[t]
      \centering
      \includegraphics[keepaspectratio, width=3.2in]{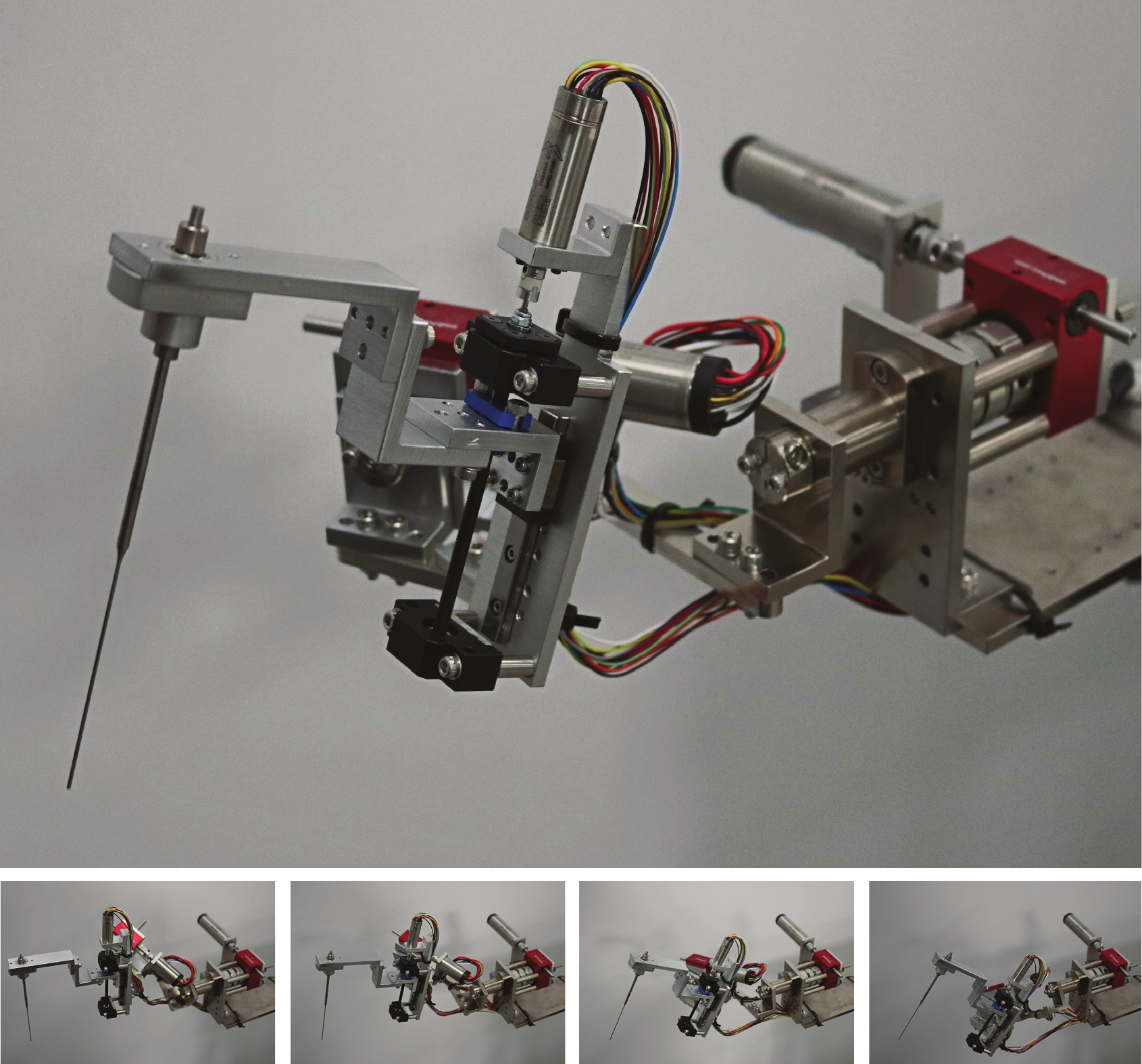}
      \caption{A 4-DoF robotic tool for vitreoretinal surgery. Bottom row: various orientations of the mechanism.}
      \label{fig:vrs_tool}
      \vspace{-0.4cm}
\end{figure}
In particular, this work contributes an integrated kinematics- and dynamics-based approach for the analysis and engineering of a self-locking 4-DoF SSM. We begin by presenting a geometric approach using twist parametrization, which yields a compact \textit{analytical} workspace representation that explicitly relates joint-axis geometry to the achievable orientation workspace. This approach enables the selection of rotation axis angles based on specific motion requirements, which can be more challenging for optimization-based methods to achieve with the same speed and intuitiveness. The analysis is further complemented by a geometrically meaningful, fast-to-compute, and numerically stable closed-form solution for the inverse kinematics, utilizing the Paden-Kahan subproblems~\cite{mls}. 

Next, we analyze the SSM when driven by high-impedance transmissions, where actuator behavior is dominated by friction. This transmission-aware dynamics modeling informs motor selection via torque prediction for low velocity surgical tasks. Using the dynamics detailed in~\cite{mablekos1, mablekos2, mablekos3, mablekos_thesis}, we release an open-source software package~\cite{mablekos_online} to calculate the robot's power requirements and estimate the friction parameters of the transmission systems, both essential for proper motorization and the development of accurate equations of motion (EoM).
\begin{figure*}[t]
    \centering
    \includegraphics[keepaspectratio,width=\textwidth]{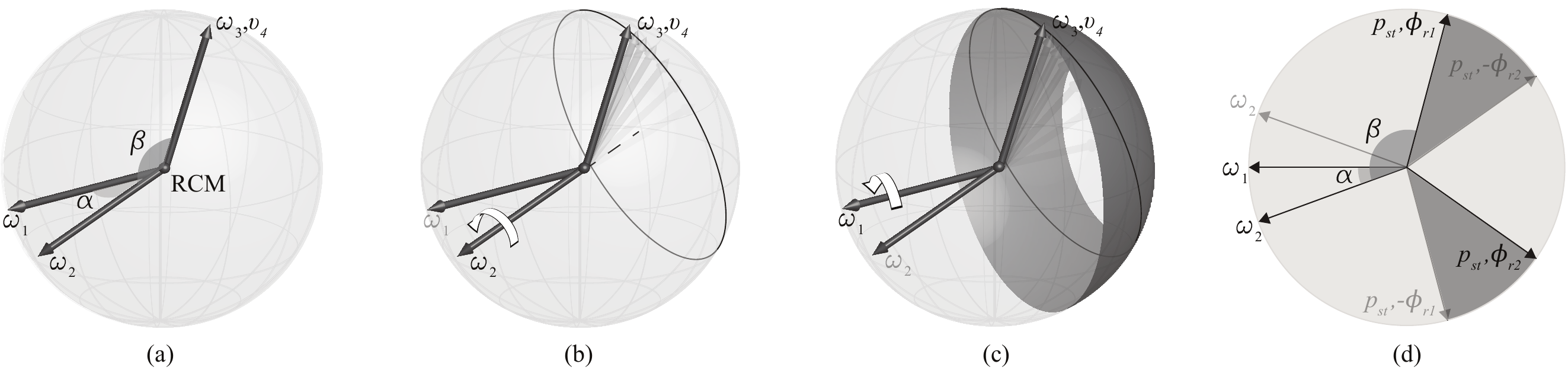}
    \caption{Workspace calculations of the 4-DoF SSM: (a) numbering conventions and twist parametrization, (b) rotation of $v_4$ about $\omega_2$, (c) rotation of ${e^{\hat{\omega}_2\theta_2}v_4}$ about $\omega_1$, (d) tilt angle range.}
    \label{fig:workspace}
    \vspace{-0.2cm}
\end{figure*}

We demonstrate the utility of the proposed framework by engineering a robotic tool for vitreoretinal surgery (see Fig.~\ref{fig:vrs_tool}). The presented kinematics and dynamics are applied to satisfy the workspace and actuation requirements of the application, while experimental results confirm the predictive capability of the analytical models, demonstrating $98\%$ agreement between theoretical and measured workspace limits and over $85\%$ agreement between simulated and measured torque responses.

\section{Kinematics} 
\label{sec:kinematics}

This section discusses the kinematics of the 4-DoF SSM for the general case of arbitrary angles between the rotation axes. We begin by introducing some preliminaries and the notation required for the kinematics description.

\subsection{Parametrization \& Forward Kinematics Preliminaries}
\label{sec:kinematics_fwd_kinematics}

Consider the 4-DoF SSM comprising of three rotational joints and one translational unit that moves the end-effector along a line passing through the RCM. We place the base frame at the RCM and the tool frame at the end-effector with initial configuration ${g_0 = (R_0,0) \in SE(3)}$. Joints $1$ to $4$ are connected with links such that joint $i$ connects links ${i-1}$ and $i$. The rotational joints are numbered from $1$ to $3$ with directions $\omega_1$, $\omega_2$, and ${\omega_3}$, respectively. The translational unit corresponds to joint~$4$ with direction $v_4$, which coincides with ${\omega_3}$, as shown in Fig.~\ref{fig:workspace}(a). The angle between $\omega_1$ and $\omega_2$ is denoted by $\alpha$, while the angle between $\omega_2$ and ${\omega_3}$ by $\beta$. Then, the unit twist coordinates ${\xi_i \in \mathbb{R}^6}$ for the $i$th joint can be written as
\begin{equation}
    \xi_i =
        \begin{cases}
            (0,\omega_i) \textnormal{ } i=1,2,3, \\
            (v_i,0) \textnormal{ } i=4. \\
        \end{cases}
    \label{eq:twists_coords}
\end{equation}

Assuming unrestricted motion for any joint, the joint space $Q$ for the 4-DoF SSM is ${Q = \mathbb{T}^3 \times \mathbb{R}}$, where $\mathbb{T}^3$ represents the $3$-torus. Given a set of joint angles ${\theta = (\theta_1,\theta_2,\theta_3,\theta_4) \in Q}$, the tool frame relative to the base frame is represented by a mapping ${g_{st}:Q \mapsto SE(3)}$. Utilizing the product of exponentials~\cite{mls}, the SSM forward kinematics can be written as ${g_{st}(\theta) = (R_{st}, p_{st})}$ with
\begin{subequations}
\label{eq:SSM_forward_kinematics}
    \begin{align}
        &R_{st} =e^{\hat{\omega}_1\theta_1} e^{\hat{\omega}_2\theta_2} e^{\hat{\omega}_3\theta_3} R_0, \label{eq:SSM_orientation} \\
        &p_{st} =e^{\hat{\omega}_1\theta_1} e^{\hat{\omega}_2\theta_2} v_4 \theta_4. \label{eq:SSM_position_vector}
    \end{align}
\end{subequations}

\subsection{Geometric Workspace Analysis}
\label{sec:kinematics_workspace}

We define the roll axis to coincide with the axis of $\omega_1$, and the tilt axis, being perpendicular to $\omega_1$, to lie in the plane spanned by $\omega_1$ and $\omega_2$. To analyze the pivoting motion generated by the SSM, we consider a geometric representation of its workspace about the RCM, capturing the achievable angular range. This range, defined by rotations about the roll and tilt axes, can be described as the set of reachable points on the unit sphere centered at the RCM, from which the achievable angular range can be inferred. Thus, we define the SSM workspace as
\begin{equation}
    W := \set[\big]{ p_{st} \given \theta_1, \theta_2 \in \mathbb{S}^1, \theta_4 = 1} \subseteq \mathbb{S}^2,
    \label{eq:SSM_workspace}
\end{equation}
where $p_{st}$ is given by~(\ref{eq:SSM_position_vector}) and $\mathbb{S}^2$ denotes the $2$-sphere.

Observing~(\ref{eq:SSM_position_vector}) for the conditions of~(\ref{eq:SSM_workspace}), the SSM workspace can be visualized as follows. First, since ${e^{\hat{\omega} \theta}}$ corresponds to a rotation about $\omega$ by $\theta$, the product ${e^{\hat{\omega}_2\theta_2} v_4}$ for ${\theta_2 \in \mathbb{S}^1}$ generates a circle on the unit sphere with radius ${|| v_4 - \omega_2 \omega_2^T v_4 ||}$ and its center lying on the $\omega_2$ axis [see Fig.~\ref{fig:workspace}(b)]. Then, the product $e^{\hat{\omega}_1\theta_1} (e^{\hat{\omega}_2\theta_2} v_4)$ for ${\theta_1 \in \mathbb{S}^1}$, which forms $W$, generates a family of concentric circles on the unit sphere with radii ${||e^{\hat{\omega}_2\theta_2} v_4 - \omega_1 \omega_1^T e^{\hat{\omega}_2\theta_2} v_4 ||}$ and centers lying on the $\omega_1$ axis [see Fig.~\ref{fig:workspace}(c)].

Considering that $W$ results from unrestricted rotation about the roll axis $\omega_1$, the SSM workspace can be fully characterized by its tilt angle range. Specifically, the intersection of $W$ with any plane containing the RCM and $\omega_1$ forms arcs whose angular extent and position relative to $\omega_1$ determine the achievable tilt range. These quantities can be extracted by examining the dot product of $\omega_1$ and $p_{st}(\theta)$ under the conditions of~(\ref{eq:SSM_workspace}).

Let ${f:\mathbb{S}^1 \mapsto [-1,1]}$ be the dot product function defined by the equation
\begin{equation}
    \label{eq:workspace_dot_product}
    \begin{aligned}
        f(\theta_2) &:= \omega_1^T (e^{\hat{\omega}_1\theta_1} e^{\hat{\omega}_2\theta_2} v_4) =  (e^{-\hat{\omega}_1\theta_1} \omega_1)^T e^{\hat{\omega}_2\theta_2} v_4 \\
                    &= \omega_1^T e^{\hat{\omega}_2\theta_2} v_4 = \cos \phi,
    \end{aligned}
\end{equation}
where $\phi$ denotes the angle between $\omega_1$ and $p_{st}(\theta)$. The first and second derivatives of $f$ with respect to $\theta_2$ are
\begin{subequations}
    \label{derivatives}
    \begin{align}
        \frac{\partial f}{\partial \theta_2} &= \omega_1^T \hat{\omega}_2 e^{\hat{\omega}_2\theta_2} v_4 =  \omega_1^T (\omega_2 \times e^{\hat{\omega}_2\theta_2} v_4) \nonumber \\
                                             &= (\omega_1 \times \omega_2)^T (e^{\hat{\omega}_2\theta_2} v_4),
                                             \label{first_derivative} \\
        \frac{\partial^2 f}{\partial \theta_2^2} &= \omega_1^T \hat{\omega}_2^2 e^{\hat{\omega}_2\theta_2} v_4. \label{sec_derivative}
    \end{align}
\end{subequations}

We examine the derivatives of (\ref{eq:workspace_dot_product}) to determine the extreme values of the angle $\phi$. According to (\ref{first_derivative}), solving ${\frac{\partial f}{\partial \theta_2}=0}$ for $\theta_2$ implies that $f(\theta_2)$ and, consequently, ${\cos \phi}$ take a potential extreme value when ${e^{\hat{\omega}_2\theta_2} v_4}$ is perpendicular to ${\omega_1 \times \omega_2}$, which implies that it lies in the plane defined by $\omega_1$ and $\omega_2$ and, therefore, can be represented as their linear combination. Denoting with $\theta_r$ the amount of rotation for which the critical points of $f$ occur and expressing ${e^{\hat{\omega}_2\theta_r} v_4}$ as a linear combination of $\omega_1$ and $\omega_2$ yields
\begin{subequations}
    \begin{align}
        &e^{\hat{\omega}_2\theta_r} v_4 = c_1 \omega_1 + c_2 \omega_2 \implies \label{eq:lin_combination} \\
        &c_2 = \omega_2^T v_4 - c_1\omega_1^T\omega_2, \label{eq:c1c2_first}
    \end{align}
\end{subequations}
where $c_1$, ${c_2 \in \mathbb{R}}$. Also, considering that a rotation preserves vector length and, thus, ${|| e^{\hat{\omega}_2\theta_2} v_4 || = || v_4 ||}$, one can write
\begin{equation}
    \label{eq:c1c2_second}
        || c_1 \omega_1 + c_2 \omega_2 || = 1.
\end{equation}
Solving (\ref{eq:c1c2_first}) and (\ref{eq:c1c2_second}) for $c_1$ and $c_2$ and substituting to (\ref{eq:lin_combination}) yields
\begin{equation}
    \label{eq:c1c2}
        e^{\hat{\omega}_2\theta_r} v_4  = \pm \frac{\sin{\beta}}{\sin{\alpha}} \omega_1 + \frac{\sin{(\alpha \mp \beta)}}{\sin{\alpha}} \omega_2.
\end{equation}
Substituting the expression for ${e^{\hat{\omega}_2\theta_r} v_4}$ to (\ref{sec_derivative}) yields a nonzero second derivative for any ${\alpha, \beta \neq 0}$ implying that ${\cos \phi}$ indeed takes an extreme value ${\cos \phi_r}$ in case ${\frac{\partial f}{\partial \theta_2}=0}$. Thus, combining (\ref{eq:workspace_dot_product}) and (\ref{eq:c1c2}) gives ${f(\theta_r) = \cos\phi_r = \cos{(\alpha \pm \beta)}}$ or, equivalently,
\begin{equation}
    \label{eq:tilt_range}
        \phi_r = \pm (\alpha \pm \beta).
\end{equation}

The four $\phi_r$ values, given by the algebraic expression in (\ref{eq:tilt_range}), correspond to the extreme angles relative to $\omega_1$ that the manipulator can achieve, thus defining the limits of the SSM workspace. These limits are graphically highlighted in Fig.~\ref{fig:workspace}(d) by the gray disk areas, which are symmetric about the $\omega_1$ axis and span an angle range of $2\alpha$ each.

The closed-form expression in (\ref{eq:tilt_range}) directly relates the achievable tilt angles to the geometric parameters $\alpha$ and $\beta$, providing an analytical characterization of the workspace limits that can be readily used for mechanism design.

\subsection{Inverse Kinematics}
\label{sec:kinematics_inv_kinematics}

Using the solutions to the Paden-Kahan Subproblems 1 and 2, explicated in~\cite{mls}, and a variation of Subproblem 3, which is analyzed in Appendix~\ref{sec:subproblem3'}, we present the techniques for converting the complete inverse kinematics problem of the 4-DoF SSM to the appropriate subproblems. 

The equation we wish to solve is
\begin{equation*}
    g_{st}(\theta) = e^{\hat{\xi}_1\theta_1} \dots e^{\hat{\xi}_4\theta_4} g_0 = g_d,
\end{equation*}
where ${g_d \in SE(3)}$ is the desired configuration of the tool frame. Post-multiplying the above equation by $g_0^{-1}$ isolates the product of exponentials
\begin{equation}
    \label{eq:exponentials_product}
        e^{\hat{\xi}_1\theta_1} \dots e^{\hat{\xi}_4\theta_4} = g_d g_0^{-1} := g_1.
\end{equation}
We determine the requisite joint angles following the three steps below.

\subsubsection[]{Solve for $\theta_4$}
Applying both sides of (\ref{eq:exponentials_product}) to a point $p_1$ that is on the axis of $\xi_4$ gives
\begin{equation}
    \label{eq:pk_step_1.1}
        e^{\hat{\xi}_1\theta_1} e^{\hat{\xi}_2\theta_2} e^{\hat{\xi}_3\theta_3} e^{\hat{\xi}_4\theta_4} p_1 = g_1 p_1.
\end{equation}
Since ${e^{\hat{\xi} \theta} q = q}$, if $q$ is on the axis of a revolute twist $\xi$, subtracting from both sides of (\ref{eq:pk_step_1.1}) a point $q_1$ that is at the intersection of $\xi_1$, $\xi_2$, and $\xi_3$ yields
\begin{equation}
    \label{eq:pk_step_1.2}
        e^{\hat{\xi}_1\theta_1} e^{\hat{\xi}_2\theta_2} e^{\hat{\xi}_3\theta_3} \big( e^{\hat{\xi}_4\theta_4} p_1 - q_1 \big) = g_1 p_1 - q_1.
\end{equation}
Taking the magnitude of both sides of (\ref{eq:pk_step_1.2}) and using the property that the distance between points is preserved by rigid motions gives
\begin{equation}
    \label{eq:pk_step_1.3}
        || e^{\hat{\xi}_4\theta_4} p_1 - q_1 || = || g_1 p_1 - q_1 ||.
\end{equation}
Equation~(\ref{eq:pk_step_1.3}) is in the form required for Subproblem~$3'$ (see Appendix), with ${p = p_1}$, ${q = q_1}$, and ${\delta = || g_1 p_1 - q_1||}$. Thus, application of Subproblem~$3'$ gives the value of $\theta_4$.

\subsubsection[]{Solve for $\theta_1$, $\theta_2$}
Rearranging (\ref{eq:exponentials_product}) by shifting the known $\theta_4$ to the right-hand side and applying both sides to a point $p_2$, which is on the axis of $\xi_3$ but not on the $\xi_1$, $\xi_2$ axes, gives
\begin{equation}
    \label{eq:pk_step_2.1}
        e^{\hat{\xi}_1\theta_1} e^{\hat{\xi}_2\theta_2} e^{\hat{\xi}_3\theta_3} p_2 = e^{\hat{\xi}_1\theta_1} e^{\hat{\xi}_2\theta_2} p_2 = g_d g_0^{-1} e^{-\hat{\xi}_4\theta_4} p_2 := q_2.
\end{equation}
Solving (\ref{eq:pk_step_2.1}) using Subproblem~$2$ for ${p = p_2}$ and ${q = q_2}$ gives the values for $\theta_1$ and $\theta_2$.

\subsubsection[]{Solve for $\theta_3$}
Isolating the unknown angle $\theta_3$ and applying both sides of the resulting kinematics to any point $p_3$, which is not on the axis of $\xi_3$, yields
\begin{equation}
    \label{eq:pk_step_2.2}
        e^{\hat{\xi}_3\theta_3} p_3 = e^{-\hat{\xi}_2\theta_2} e^{-\hat{\xi}_1\theta_1} g_d g_0^{-1} e^{-\hat{\xi}_4\theta_4} := q_3.
\end{equation}
Solving (\ref{eq:pk_step_2.2}) using Subproblem~$1$ for ${p = p_3}$ and ${q = q_3}$ gives the value of $\theta_3$, completing the solution to the inverse kinematics problem of the 4-DoF SSM.

\section{Dynamics} 
\label{sec:dynamics}

For the purposes of this paper, we focus solely on the actuator-reflected dynamics of each link of the 4-DoF SSM driven by a self-locking transmission system. These dynamics can be expressed using the typical torque formula as
\begin{equation}
    \label{eq:ssm_link_eom}
        \mathscr{J}_i \dot{w}_i = \mathscr{T}_i,
\end{equation}
where $i$ denotes the link number, $w_i$ the $i$th joint velocity, $\mathscr{J}_i$ the reflected inertia, and $\mathscr{T}_i$  the equivalent torque. The reflected inertia and equivalent torque depend on several factors, including the static $\mu_s$ and Coulomb $\mu_c$ friction between the sliding elements of the transmission, as well as the viscous $b_v$ and Coulomb $b_c$ friction of the support bearings. The software developed for this paper will later identify these parameters and use them for the inverse dynamics evaluation. The extended theory on self-locking transmission systems and the detailed reasoning behind (\ref{eq:ssm_link_eom}) can be found in~\cite{mablekos3}.

\section{Robot Design} 
\label{sec:robot_design}

In this section, we present the design of a robotic tool for vitreoretinal surgery, leveraging the previously developed workspace analysis to determine the rotation-axis angles and design the robot frame to meet the required range of motion. We then define the velocity requirements of the application and, using inverse dynamics, calculate the torque demands to guide actuator selection. Details on frame design and actuation choices are provided in the following subsections.

\subsection{Task Description \& Motion Requirements}
\label{sec:description}

Vitreoretinal surgery involves accessing the posterior cavity of the eye through the cannulas of trocar devices, which serve as fulcrum points. Robotic systems for this procedure aim to enhance interventional precision, particularly for delicate tasks such as retinal therapy delivery~\cite{vanderpoorten2020}.

\begin{figure}[t]
      \centering
      \includegraphics[keepaspectratio, width=3.2in]{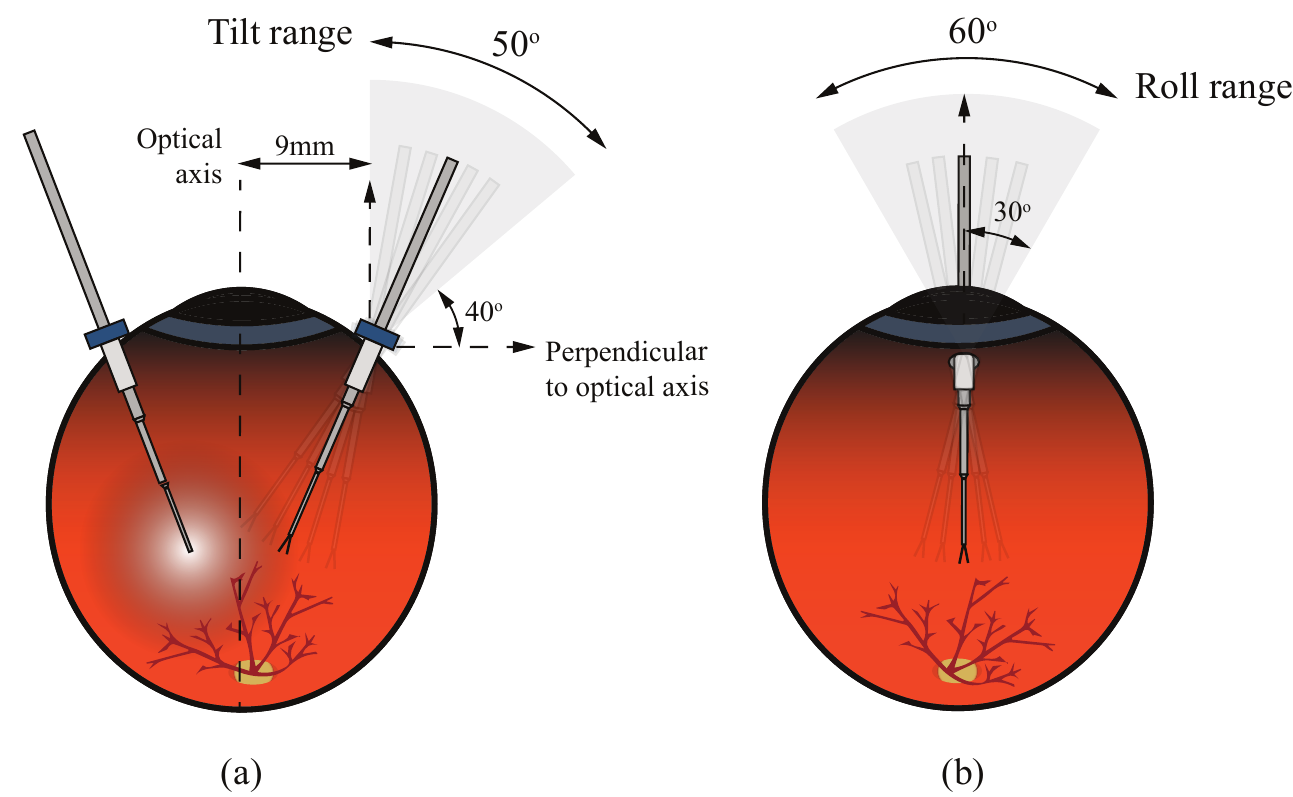}
      \caption{Eyeball plane sections and required angle range of the tool about: (a) the tilt axis, (b) the roll axis.}
      \label{fig:ranges}
      \vspace{-0.2cm}
\end{figure}

In this work, the robotic tool must provide four DoF to achieve the required motions: axial translation through the entry point, and spin, roll, and tilt about the RCM. Clinical literature~\cite{gijbels1,steadyhand1} indicates that motion ranges of $360^\circ$ spin, $60^\circ$ roll, $50^\circ$ tilt, and $30$mm translation are sufficient for surgeons to manipulate the tool effectively (see Fig.~\ref{fig:ranges}). These motion ranges directly inform the workspace analysis and guide the selection of rotation axes and link dimensions for the robot frame, ensuring practical feasibility of the design.

Note that the initial placement of the tool tip at the eye entry point is typically performed by a separate manipulator and requires at least three translational DoF. While outside the scope of this work, solutions for this phase can be found in relevant literature~\cite{smits2019, henry2025, inagaki2025}.

\subsection{Frame Design}
\label{sec:frame_design}

The structure of the robot is determined primarily by the angles between the rotation axes of the SSM, which are set based on the workspace analysis presented in Sec.~\ref{sec:kinematics}.  

\begin{figure}[t]
      \centering
      \includegraphics[keepaspectratio, width=3.3in]{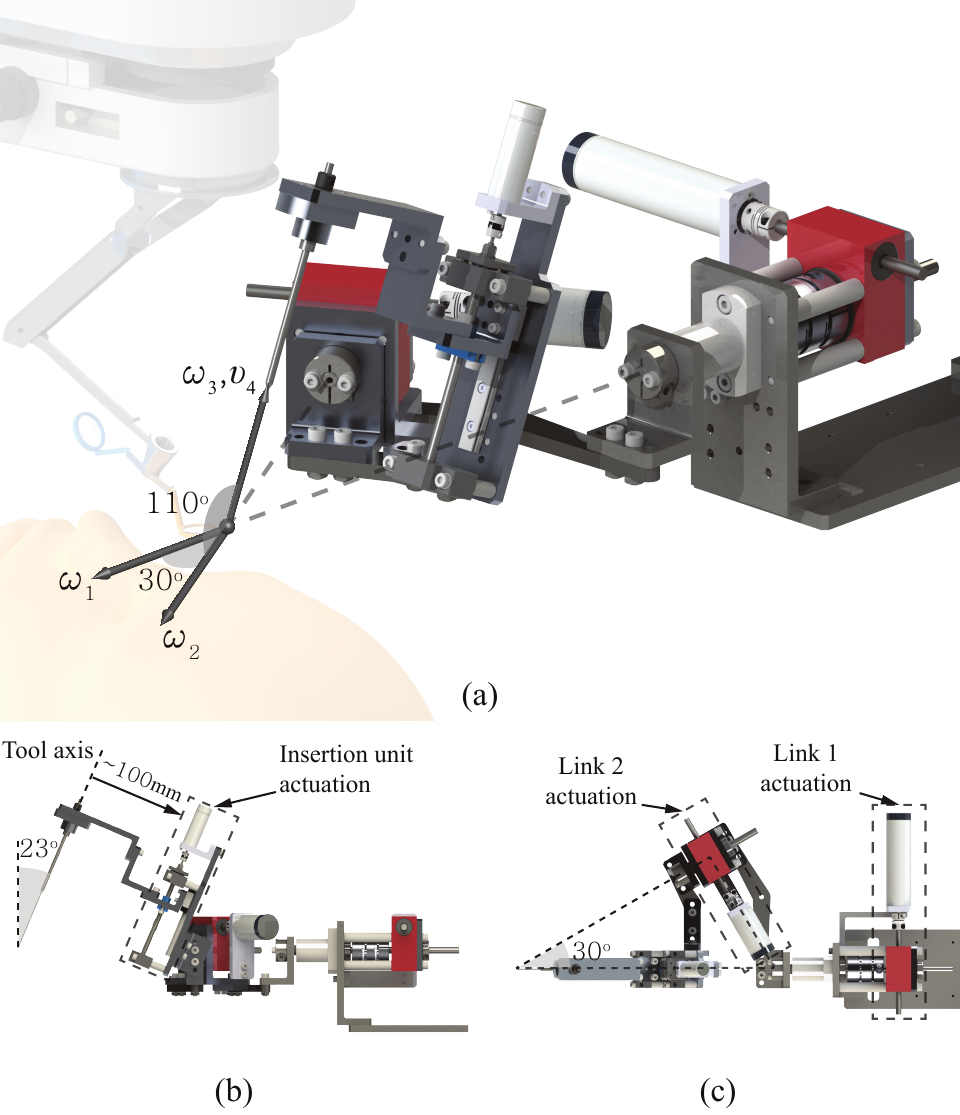}
      \caption{Vitreoretinal surgery robot CAD design: (a) isometric view, (b) side view, (c) top view. }
      \label{fig:s-robot}
      \vspace{-0.2cm}
\end{figure}

First, the angle $\alpha$ between $\omega_1$ and $\omega_2$ is set to $30^\circ$ [see Fig.~\ref{fig:s-robot}(a)], balancing the required tilt range with mechanical compactness. Next, starting from the perpendicular to the plane defined by $\omega_1$ and $\omega_2$, the initial tilt angle of $\omega_3$ is set to $-23^\circ$, resulting in an angle $\beta \approx 110^\circ$ between $\omega_2$ and $\omega_3$ [see Fig.~\ref{fig:s-robot}(a,b)]. According to (\ref{eq:tilt_range}), this configuration yields a tilt range of $60^\circ$, from $-140^\circ$ to $-80^\circ$ relative to $\omega_1$, which covers the clinically relevant motions. Because the 4-DoF SSM allows unrestricted rotation about the roll and spin axes, these chosen rotation angles satisfy the pivoting requirements about the RCM while maintaining mechanical simplicity.

Regarding the translation range, meeting the $30$mm requirement poses no difficulty. Notably, the translation unit creates a gap between the tool tip and the actuated parts of the robot, a key advantage of spherical mechanisms. This distance depends on the angles between the rotation axes and the lengths of links 1 and 2 [see Fig.~\ref{fig:s-robot}(b,c)]. With $\alpha = 30^\circ$, we achieve approximately $100$mm of separation while keeping link sizes minimal, preventing unnecessary torque increases, as calculated in Sec.~\ref{sec:actuation_selection}.

\subsection{Actuation Selection}
\label{sec:actuation_selection}

The mechanical design of the robot is finalized by selecting actuators and their transmission systems, guided by the torque requirements estimated from the inverse dynamics simulations. 

\subsubsection{Motion Transmission}

Joints 1 and 2 are driven by fully assembled worm gearboxes (Ondrives P30-120) with a high 120:1 reduction ratio, ensuring non-backdrivability and safe, self-locking operation. The translation unit uses a simple lead screw drive (Helix Linear NFA016048RS), also non-backdrivable. These choices were determined primarily by the availability of compact, self-locking transmission systems and by the estimated torque requirements based on the manipulator linkages.

\subsubsection{Motorization}

Motor selection is based on the simulated joint torques and velocities obtained from the inverse dynamics of (\ref{eq:ssm_link_eom}), using the CAD-derived inertial parameters and manufacturer-provided friction data for each transmission. The maximum accelerations and velocities for the motorized rotating joints are approximately $200^\circ$/s$^2$ and $65^\circ$/s, while the translation unit requires $100$mm/s$^2$ acceleration and $20$mm/s velocity. These values, derived from literature~\cite{nasseri1}, ensure the robot meets the clinical motion requirements.  

\begin{figure}[t]
      \centering
      \includegraphics[keepaspectratio, width=3.1in]{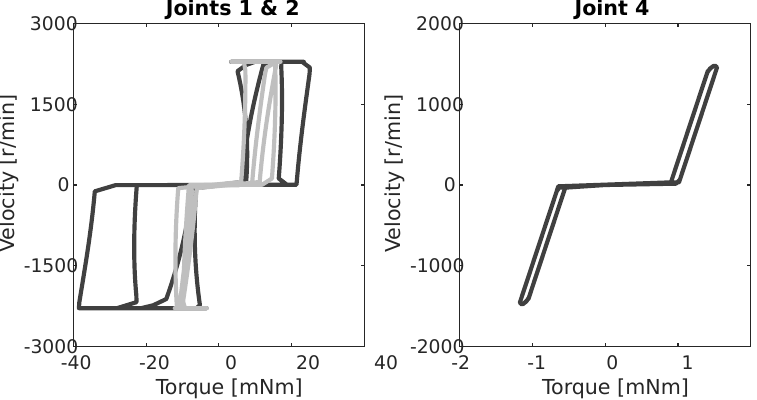}
      \vspace{-0.1cm}
      \caption{Actuator reflected payload curves: joint 1 (left subfigure - gray~line), joint 2 (left subfigure - pale gray line), joint 4 (right subfigure).}
      \label{fig:payload_curves}
      \vspace{-0.2cm}
\end{figure}

The torque-velocity load curves generated by the simulation (see Fig.~\ref{fig:payload_curves}) guide the selection of appropriate motors (Maxon ECXTQ22XL, ECXTQ22M, ECXSP13M). All scripts used to model the 4-DoF SSM, configure its transmissions, and simulate motion responses are available online~\cite{mablekos_online}. This workflow enables reproducible design iterations, allowing designers to explore alternative transmissions, motor sizes, or motion requirements efficiently.

\section{Experiments} 
\label{sec:experiments}

This section presents experiments conducted on the developed prototype to validate the proposed framework and demonstrate its practical utility for microsurgical tool design. The experiments aim to (i) verify the analytical workspace predictions and (ii) provide experimental data for friction parameter identification, essential for accurate torque estimation and actuator selection. For all experiments, motor velocities are commanded via closed-loop control, and the motor torque and velocity are recorded at 200 Hz. These signals form the basis for both workspace evaluation and friction parameter estimation.

\subsection{Workspace Verification}
\label{sec:workspace_verification}

\begin{figure}[t]
      \centering
      \includegraphics[keepaspectratio, width=\columnwidth]{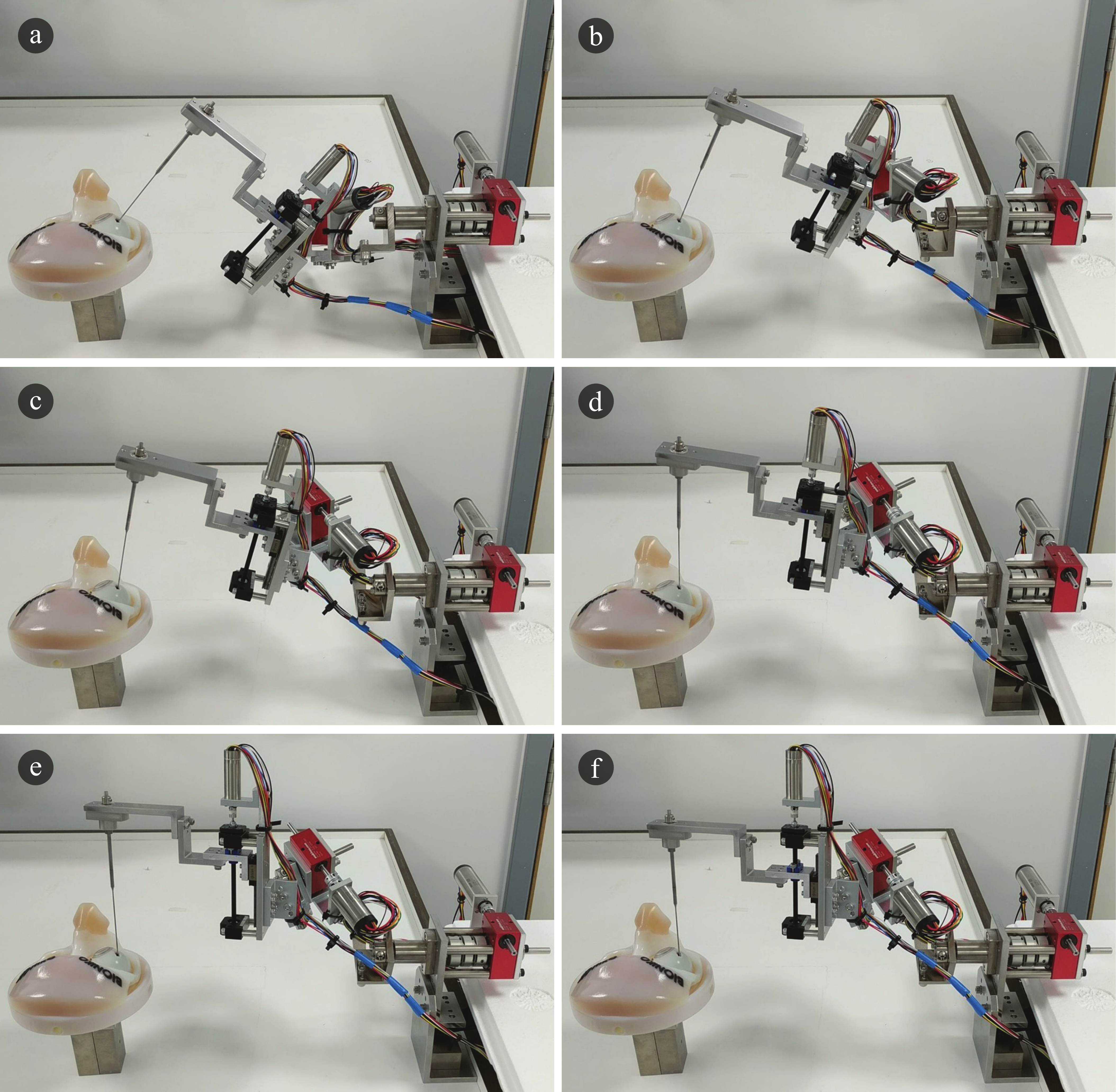}
      \vspace{-0.5cm}
      \caption{Video snapshots of the tool rotating about the tilt axis (a-e) followed by tool insertion (e-f).}
      \label{fig:snaps}
      \vspace{-0.0cm}
\end{figure}

To verify the analytical workspace, the tool tip is placed at the RCM, and the rotation joints are moved to the predicted workspace limits using the inverse kinematics solution. The actual tilt angles are measured with a universal protractor relative to the horizontal reference plane of the robot’s base. Video snapshots are captured to illustrate the tool’s motion (Fig.~\ref{fig:snaps}).

The measured tilt ranges are compared with the theoretical values derived from (\ref{eq:tilt_range}) for the design described in Sec.~\ref{sec:frame_design}. The results, reported in Table~\ref{table:kin_eval}, show a $98.5\%$ agreement between the theoretical and measured tilt ranges. Minor discrepancies are attributed to manufacturing and assembly tolerances. These results confirm that the proposed geometric framework reliably predicts the workspace, providing designers with accurate information for selecting rotation axes and link dimensions to meet application-specific motion requirements.

\begin{table}[t]
    \centering
    \caption{Theoretical vs Measured Tilt Range}
    \vspace{-0.1cm}
    {\setlength{\extrarowheight}{1pt}
		\begin{tabular}{r c cc} 
    	    \hline
    	    \hline
    	    \addlinespace[2pt]
    	    \textbf{Parameter} && \textbf{Theoretical} & \textbf{Measured} \\
    	    \cline{1-1} \cline{3-4}
    	    \addlinespace[2pt]
    	
        	Tilt Range && 60$^\circ$ & 59.1$^\circ$ \\
            \makecell[r]{Range w.r.t.\\$\omega_1$-axis} && -140$^\circ$ to -80$^\circ$ & -138.6$^\circ$ to -79.5$^\circ$ \\
            \hline
    	\end{tabular}}
    \label{table:kin_eval}
\end{table}

\subsection{Dynamics Identification \& Evaluation}
\label{sec:dyn_id}

This subsection describes the identification of joint friction parameters and the evaluation of the inverse dynamics model for the 4-DoF SSM. These parameters are critical for predicting actuator torques, which guide motor selection and inform design trade-offs.

\paragraph{Identification}

\begin{figure}[t]
      \centering
      \includegraphics[keepaspectratio, width=\columnwidth]{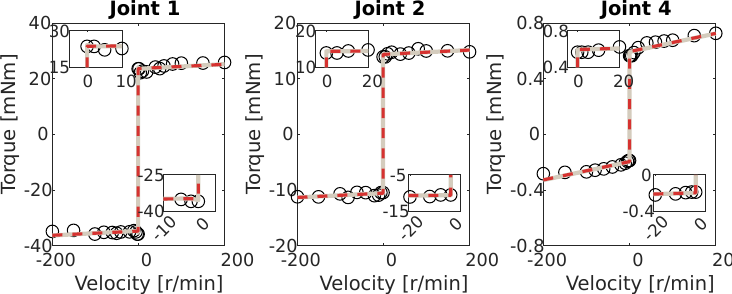}
      \vspace{-0.5cm}
      \caption{Experimental (black circles) and simulation (dashed red line) torque-velocity maps for the actuated joints of the prototype.}
      \label{fig:ssm_maps}
      \vspace{-0.0cm}
\end{figure}

Each joint is tested independently, with no external loads. Constant-velocity motions are commanded, and the resulting torque-velocity data are fitted to a static-plus-kinetic friction model (Sec.~\ref{sec:dynamics}), capturing the dominant effect of the high-impedance, self-locking transmissions. The torque-velocity curves are constructed for each joint following the single-DoF method in~\cite{mablekos3, mablekos_thesis}. The software package accompanying this paper~\cite{mablekos_online} provides the scripts to generate the curves and fit the friction parameters. The measured torque-velocity curves, along with the fitted model, are shown in Fig.~\ref{fig:ssm_maps}, and the estimated friction parameters are summarized in Table~\ref{table:ssm_estimated_params}.

\begin{table}[t]
    \centering
    \caption{Friction Parameters in SI}
    \vspace{-0.1cm}
    {\setlength{\extrarowheight}{1pt}
		\begin{tabular}{r c c c c c c} 
    	    \hline
    	    \hline
        	\textbf{\makecell[r]{Parameter}} &&
                \textbf{\makecell{Joint 1}} &&
                \textbf{\makecell{Joint 2}} &&
                \textbf{\makecell{Joint 4}} \\
                \cline{1-1} \cline{3-7}
                \addlinespace[2pt]
    
                $\mu_c$ && $0.13$ && $0.12$ && $0.17$ \\
    
                $\mu_{s}$ && $0.15$ && $0.13$ && $0.17$ \\
        	
                $10^3b_c$ && $3.82$ && $3.54$ && $0.11$ \\
            
                $10^5b_v$ && $7.18$ && $3.69$ && $0.62$ \\
    
                \hline
    	\end{tabular}}
    \label{table:ssm_estimated_params}
\end{table}

\paragraph{Evaluation}

The identified friction parameters are used to simulate representative tool motions relevant to vitreoretinal surgery, including combined tilt rotations and axial insertion (motion similar to the video snapshots shown in Fig.~\ref{fig:snaps}). Tilt rotation is chosen, as it requires combined motion from all the rotational joints. Experimental joint velocities serve as inputs to the inverse dynamics simulation. The predicted torque responses are compared against experimental measurements using the normalized RMSD (NRMSD) metric \cite{mablekos3}. Table~\ref{table:ssm_evaluation} reports NRMSD values below 13\% for all joints, demonstrating that the model accurately predicts actuator torques for the motions of interest. This confirms the practical utility of the framework: designers can rely on the simulation to estimate torque requirements and select appropriate motors and transmission systems without extensive experimental testing.

\begin{table}[h]
    \caption{Simulated vs Experimental Torque NRMSD}
    \vspace{-0.1cm}
    \centering
    {\setlength{\extrarowheight}{1pt}
		\begin{tabular}{r c ccc} 
    	    \hline
    	    \hline
    	    \addlinespace[2pt]
    	    \textbf{\makecell[c]{Speed\\(Tilt-Insertion)}} && \textbf{Joint 1} & \textbf{Joint 2} & \textbf{Joint 4} \\
    	    \cline{1-1} \cline{3-5}
    	    \addlinespace[2pt]
    	
            10$^\circ$/s - 10mm/s && 0.08 & 0.12 & 0.12 \\
    	    20$^\circ$/s - 20mm/s && 0.11 & 0.10 & 0.13 \\
    	    40$^\circ$/s - 40mm/s && 0.11 & 0.11 & 0.09 \\
            80$^\circ$/s - 40mm/s && 0.09 & 0.12 & 0.10 \\
            \hline
    	\end{tabular}}
    \label{table:ssm_evaluation}
\end{table}
%

\section{Conclusion}
\label{sec:conclusion}

This study presented a framework for the kinematic and transmission-aware design of a 4-DoF serial spherical mechanism for microsurgery. By combining an analytical workspace representation with friction-informed torque estimation, the approach provides a practical and reproducible methodology to select rotation axis angles, evaluate actuator requirements, and ensure desired motion capabilities. The kinematic and dynamic components are modular, allowing robotics engineers to apply either or both aspects depending on their specific design needs. While demonstrated for vitreoretinal surgery, the framework is general and can be applied to other serial spherical manipulators where precise motion, workspace adjustment, and/or self-locking operation are critical. Future work could explore the integration of more advanced kinematic metrics or optimization-driven design refinements to further streamline the engineering workflow.




\appendices

\section{Subproblem \texorpdfstring{$3'$}~~: Translation to a Given Distance}
\label{sec:subproblem3'}

In this appendix, we provide a solution to the geometric problem that occurred in the first step of simplification of the 4-DoF SSM inverse kinematics. This problem corresponds to translating a point along a given axis until it is at a specific distance from another given point in space. The mathematical formulation of this subproblem is as follows. Let $\xi$ be an infinite-pitch unit-magnitude twist; $p$, ${q \in \mathbb{R}^3}$ two points; and $\delta$ a positive real number. Find $\theta$ such that
\begin{equation}
    \label{eq:app_1}
        || q - e^{\hat{\xi} \theta} p || = \delta.
\end{equation}

To find the explicit solution, we first consider the unit vector $v$ such that ${\xi = (v,0)}$. Then, the exponential of the corresponding twist is ${e^{\hat{\xi} \theta} = (I, v\theta)}$. Equation~(\ref{eq:app_1}) can therefore be written as
\begin{equation}
    \label{eq:app_2}
        || u - v \theta || = \delta,
\end{equation}
where ${u = q - p}$. Squaring both sides of (\ref{eq:app_2}) yields
\begin{align*}
        &( u - v \theta )^T( u - v \theta ) = \delta^2 \implies \\
        &\theta^2 - 2u^T v\theta + ||u||^2 - \delta^2 = 0,
\end{align*}
and, therefore,
\begin{equation}
    \label{eq:app_3}
        \theta = u^T v \pm \sqrt{(u^T v)^2 + \delta^2 - ||u||^2}.
\end{equation}

Equation~(\ref{eq:app_3}) gives zero, one, or two real solutions for $\theta$, depending on the number of points in which the line defined by the $\xi$ axis intersects the sphere of radius $\delta$ centered at $q$.





{\bibliographystyle{IEEEtran}
\bibliography{Bibtex/bibliography}
}

\end{document}